\documentclass{article}

\usepackage{arxiv}

\usepackage[utf8]{inputenc} 
\usepackage[T1]{fontenc}    
\usepackage{hyperref}       
\usepackage{url}            
\usepackage{booktabs}       
\usepackage{amsfonts}       
\usepackage{nicefrac}       
\usepackage{microtype}      
\usepackage{lipsum}		
\usepackage{graphicx}
\usepackage{natbib}
\usepackage{doi}

\usepackage[linesnumbered,ruled,vlined]{algorithm2e}
\usepackage{mathtools}

\title{FedQAS: Privacy-aware machine reading comprehension with federated learning}

\author{Addi Ait-Mlouk \\
        Department of Information Technology\\
        Division of Scientific Computing\\
        Uppsala University, Sweden\\
        \texttt{addi.ait-mlouk@it.uu.se \thanks{Corresponding author.}}
         \And
        Sadi Alawadi\\
        Department of Information Technology\\
        Division of Scientific Computing\\
        Uppsala University, Sweden\\
        \texttt{sadi.alawadi@it.uu.se}
        \And
        Salman Toor\\
        Department of Information Technology\\
        Division of Scientific Computing\\
        Uppsala University, Sweden\\
        Scaleout Systems, Sweden\\
        \texttt{salman.toor@it.uu.se}
         \And
         Andreas Hellander \\
        Department of Information Technology\\
        Division of Scientific Computing\\
        Uppsala University, Sweden\\
        Scaleout Systems, Sweden\\
        \texttt{andreas.hellander@it.uu.se}
}

\renewcommand{\shorttitle}{Privacy-aware machine reading comprehension with federated learning}

\hypersetup{
pdftitle={A template for the arxiv style},
pdfsubject={q-bio.NC, q-bio.QM},
pdfauthor={David S.~Hippocampus, Elias D.~Striatum},
pdfkeywords={First keyword, Second keyword, More},
}

\begin{document}
\maketitle

\begin{abstract}
Machine reading comprehension (MRC) of text data is one important task in Natural Language Understanding. It is a complex NLP problem with a lot of ongoing research fueled by the release of the Stanford Question Answering Dataset (SQuAD) and Conversational Question Answering (CoQA). It is considered to be an effort to teach computers how to "understand" a text, and then to be able to answer questions about it using deep learning. However, until now large-scale training on private text data and knowledge sharing has been missing for this NLP task. Hence, we present FedQAS, a privacy-preserving machine reading system capable of leveraging large-scale private data without the need to pool those datasets in a central location. The proposed approach combines transformer models and federated learning technologies. The system is developed using the FEDn framework and deployed as a proof-of-concept alliance initiative. FedQAS is flexible, language-agnostic, and allows intuitive participation and execution of local model training. In addition, we present the architecture and implementation of the system, as well as provide a reference evaluation based on the SQUAD dataset, to showcase how it overcomes data privacy issues and enables knowledge sharing between alliance members in a Federated learning setting.
\end{abstract}


\section{Introduction}
\label{sec:introduction}
\begin{figure}
  \centering
  \includegraphics[width=8cm]{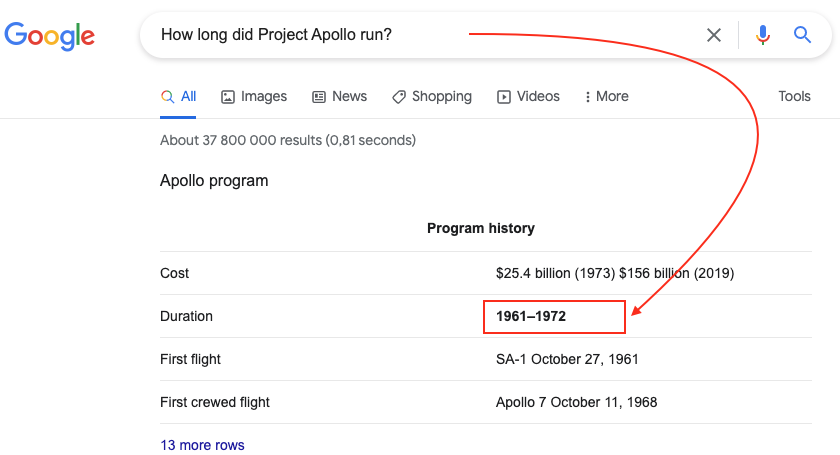}
  \caption{An example of Google search engine with machine reading comprehension techniques.}
  \label{fig1:mrc}
\end{figure}

Machine reading comprehension (MRC) is a sub-field of natural language understanding (NLU), it aims to teach machines to read and understand human languages (text). A user can ask the machine to answer questions based on a given paragraph or text document. Generally, MRC requires modeling complex interactions between the context and the query in a specific domain. It could be used in many NLP applications such as dialogue systems and search engines as shown in Figure \ref{fig1:mrc} -- a Google search engine with MRC techniques can directly return the correct answers to questions rather than a list of content and web pages. These kinds of techniques have been based on hand-crafted rules that need substantial human effort and resources. However, recently there has been an explosion of various MRC benchmark datasets that leads to a variety of models such as BiDAF \citep{seo2018bidirectional} and other models based on BERT \citep{devlin-etal-2019-bert}, RoBERTa \citep{zhuang-etal-2021-robustly}, XLNet \citep{NEURIPS2019_dc6a7e65}, ELMo \citep{peters-etal-2018-deep} and transformer \citep{vaswani2017attention}.

Despite this rapid progress on MRC datasets and models, most of the existing work has focused on algorithms for improving model performance. 
At present, several MRC models have already surpassed human performance on many of the MRC datasets \citep{rajpurkar2016squad}, but there is still a limit in terms of data privacy, collaborative training, resource-consuming, and security. Hence, there is a need of extending existing MRC models to the context of federated learning and decentralized AI for system performance, collaborative privacy-preserving training, and knowledge sharing by participating in collaborative training. To address these gaps, we proposed a privacy-aware approach based on federated learning technology to learn new global models in a geographically distributed manner using FEDn framework \citep{ekmefjord2021scalable}, build more challenging MRC models by integrating with private data generation and labeling for local accurate training as well as incremental learning approach to strengthening the model performances during collaborative training without compromising the data.\\

The remainder of this paper is organized as follows. Section \ref{sec:relatedwork} surveys related work. Section \ref{sec:approach} details the proposed approach and architecture of FedQAS, with an emphasis on its privacy and scalability properties. In Section \ref{sec:experiments}, we demonstrate the frameworks potential in an evaluation based on the SQUAD dataset. Finally, section 4  concludes the work and outlines future work.

\section{Related work}
\label{sec:relatedwork}
Machine reading comprehension was proposed for the first time in 1977 by Lehnert, who built a question answering program called the QUALM \citep{Alsolami2012Auth}. In 1999, Hirschman et al. \citep{10.3115/1034678.1034731} built a reading comprehension system using a corpus of 60 development and 60 test stories of 3rd to 6th-grade material. Because of the lack of benchmark datasets in that period, most MRC systems were rule-based or statistical models \citep{10.3115/1117595.1117598,charniak-etal-2000-reading}. In recent years, many benchmark datasets have been released that focus on MRC by answering questions, see Table \ref{dataset-tab}. Since these datasets were made available, there has been considerable progress on MRC tasks. The Stanford Question Answering Dataset (SQuAD) is one of the most well-known reading comprehension datasets, consisting of 100.000 questions posed by crowd-workers on Wikipedia articles, where the answer to every question is a segment of text, or span, from the corresponding reading context. Important progress based on SQUAD concerns include the attention method \citep{wang2017bilateral} and Bi-Directional Attention Flow (BiDAF)\citep{richardson-etal-2013-mctest} which improved considerably the question answering performance. These two methods compute Context to Question attention and Question to Context attention using a similarity matrix computed directly from context and question. Authors in \citep{wang-etal-2018-multi-granularity} describe a novel hierarchical attention network for reading comprehension style question answering, which aims to answer questions for a given narrative paragraph. In their work, attention and fusion are conducted horizontally and vertically across layers at different levels of granularity between question and paragraph. In recent work in language modeling, authors in \citep{zhang2020semanticsaware} incorporate explicit contextual semantics from pre-trained semantic role labeling and introduce an improved language representation model, Semantics-aware BERT (SemBERT), which is capable of explicitly absorbing contextual semantics over a BERT backbone. Moreover, Zhuosheng et al.\citep{zhang2019sgnet} propose using syntax to guide the text modeling by incorporating explicit syntactic constraints into the attention mechanism for better linguistically motivated word representations. Other relevant works have been proposed including \citep{yamada2020luke,lan2020albert,zhang2020retrospective}.\\

All these proposed approaches required a very large amount of data for training, which is not always available in some cases, in particular when the text data is sensitive, private (medical text, business, social media), and very big. In this context, we here propose the use of federated learning as a method for distributed and collaborative machine learning. Organizations maintain and govern their data locally and participate in learning a new global, federated model by sending only their model updates (model weights) to a server for aggregation into the global model. Hence, all participants (clients) can benefit from a newly trained model without exposing their data publicly. 

Our contributions in this paper can be summarized as follows: (1) We propose federated learning models for MRC using a transformer architecture, (2) We design and develop the FedQAS system for collaborative training, (3) we preserve data privacy (4) we improve the local training with incremental learning scheme and private data generation (5) Our analyses of the models respect data privacy regulations and outperforms the baseline model on SQuAD.

\begin{table}
  \centering
  \caption{List of some existing MRC datasets}
  \label{dataset-tab}
  \begin{tabular}{lll}
  \hline\noalign{\smallskip}
    Dataset  & Answer Type & Domain \\
    \noalign{\smallskip}\hline\noalign{\smallskip}
	MCTest \citep{richardson-etal-2013-mctest}	& Multiple choice & Children’s \\ 
	    &   & stories  \\
	CNN/Daily Mail \citep{hermann2015teaching} & Spans & News \\
	
	Children’s book \citep{hill2016goldilocks}  & Spans & Children’s \\
	&   & stories  \\
	MS MARCO \citep{bajaj2018ms} & Free-form text & Web Search \\
	NewsQA \citep{trischler-etal-2017-newsqa} & Spans & News \\
	SearchQA \citep{dunn2017searchqa} & Spans & Jeopardy \\
	TriviaQA \citep{joshi2017triviaqa} & Spans & Trivia \\
	SQuAD \citep{rajpurkar-etal-2016-squad} & Spans & Wikipedia \\
	SQuAD 2.0 \citep{rajpurkar-etal-2018-know} & Spans, Unanswerable & Wikipedia\\
	CoQA \citep{reddy-etal-2019-coqa} & Free-form text, & News, Reddit \\
	     & & Wikipedia  \\
\noalign{\smallskip}\hline
\end{tabular}
\end{table}

\section{Proposed approach}
\label{sec:approach}
\begin{figure*}
  \includegraphics[width=\linewidth]{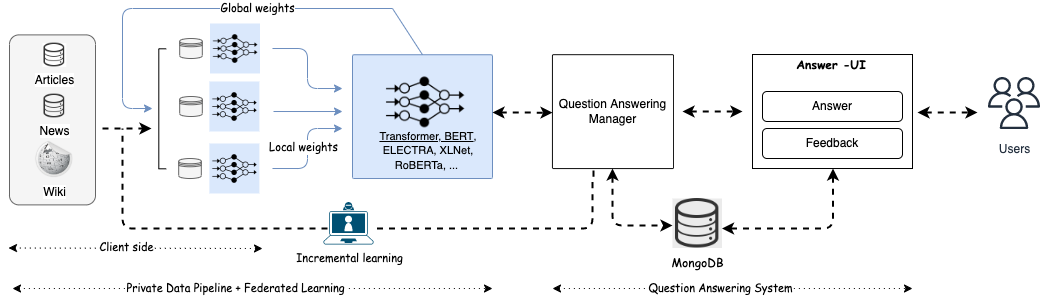}
  \caption{Overview of the FedQAS architecture. The FedQAS approach is organized in three main logical layers, the first one is for collaborative privacy-preserving training, the second one is a federated question answering manager, and the third is for incremental learning and private data generation.}
  \label{fig1:system}
\end{figure*}

The overall architecture of our proposed FedQAS system is shown in Figure \ref{fig1:system}. The main modules are private data pipeline, federated learning settings, question answering, and incremental learning. The private data pipeline module allows local users (clients) to process and prepare their data locally to be used by federated learning algorithms. The federated learning module enables multiple private clients to form an alliance to collaboratively train machine learning/deep learning models and send parameters to the server for global model generation (aggregation of local models). Afterward, the system allows the client to add new data locally and train the model incrementally through a defined number of rounds to improve the performance using incremental learning techniques. Finally, participating clients can use the global model as a question answering system. The system is implemented using the FEDn federated learning framework \citep{ekmefjord2021scalable} and Flask \footnote{https://flask.palletsprojects.com/}.FEDn provides a highly scalable federated learning run-time, and Flask is used to develop interactive and user-friendly interfaces for the different processes in the workflow. The list of available datasets related to question answering used for the demo is placed in the local data sources. Moreover, the developed system is scalable, flexible, and can be expanded with new clients/data sets on-demand (without the need to re-train the federated model).

\subsection{Data processing pipeline}
\begin{figure*}
  \includegraphics[width=\linewidth]{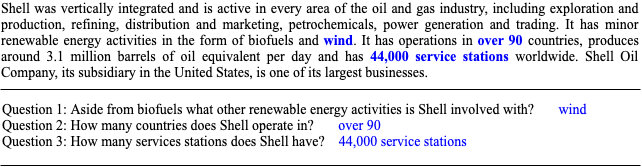}
  \caption{A paragraph from Wikipedia and three associated questions together with their answers, taken from the SQuAD dataset.}
  \label{para}
\end{figure*}

Stanford Question Answering Dataset (SQuAD) \citep{rajpurkar-etal-2018} is a machine reading comprehension dataset, consisting of questions posed by crowd-workers on a set of Wikipedia articles, where the answer to every question is a segment of text, or span, from the corresponding reading passage, alternatively the question might be unanswerable. SQuAD2.0 combines the 100.000 questions in SQuAD1.1 with over 50,000 unanswerable questions written adversarially by crowd-workers to look similar to answerable ones. To do well on SQuAD2.0, systems must not only answer questions when possible but also determine when no answer is supported by the paragraph and abstain from answering \footnote{https://rajpurkar.github.io/SQuAD-explorer/}, see Figure \ref{para} for an example of text, questions, and answers.

\subsection{Data privacy}
To train a model with a high level of accuracy, machine reading comprehension models require large datasets to ‘learn’ from, however, data might be sensitive and private. To preserve data privacy, different anonymization techniques have been used. The most relevant are k-anonymity \citep{10.1142/S0218488502001648}, l-diversity \citep{10.1145/1217299.1217302}, and t-closeness \citep{4221659}. In k-anonymity, specific columns (e.g., name, religion, sex) are removed or altered (e.g., replacing a specific age with an age span). L-diversity and t-closeness are extensions of k-anonymity, which are used to protect attribute disclosure, these anonymization techniques are applied before data is shared for training. However, with the rise of AI, this form of anonymizing personal data is not enough to protect privacy because the data can often be reverse-engineered using machine learning to re-identify individuals \citep{naturecomm}. In question answering systems, there might be sensitive documents, personal data that needs to be processed for MRC task, without exposing data. To handle this issue, we propose a question answering system based on federated learning methodology to protect data leakage and ensure secure collaborative training. The proposed system follows a federated learning paradigm in which participating clients are required to train their local models and then share the gradient (model parameters) for an eventual aggregation strategy in a central server. This approach ensures input data privacy, enables collaborative training, low-cost training by distributing the workload across clients instead of training a large model individually, sharing the local learning model within the alliance (training clients) without compromising private data, and improving local learning by using incremental learning and local data generation pipeline.

\subsection{Federated learning}
Federated learning is an emerging technology enabling multiple parties to jointly train machine learning models on private data. These parties could be mobile and IoT devices (cross-device FL), or organizations (cross-silo). Data remain locally at each party, only the parameter updates are communicated with a server and other parties. In our system, we use FL to develop FedQAS based on transformer architectures for Question Answering. FedQAS trains a global model on large amounts of data from multiple geographically distributed parties. Each party trains a local transformer model on its data (Algorithm \ref{client-update}) and sends parameters to the central server for aggregation (FedAVG \citep{mcmahan2017communicationefficient}) instead of the whole model. In the aggregation part, the aggregator (running in the \emph{combiner} in FEDn \citep{ekmefjord2021scalable}) combines parameters (Algorithm \ref{fedavg}) and generates a single global model for each round using federated incremental averaging \citep{mcmahan2017communicationefficient}. 

\subsection{Incremental Learning}
\begin{figure}[!h]
\centering
\includegraphics[width=8cm]{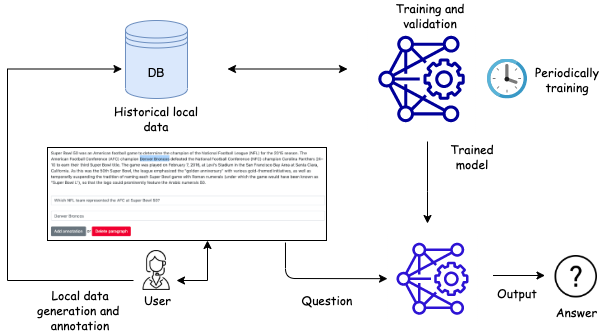}
\caption{Federated incremental learning process. Clients can add data continuously to extend the existing model’s knowledge locally while training a global model, the generated data can be stored locally on the client-side.}
\label{active_l}
\centering
\end{figure}

Incremental learning is a machine learning case in which input data is continuously used to extend the existing model's knowledge i.e. to further train the model. It attempts to improve a model’s performance while adding the fewest samples possible. In the proposed system, adding data locally by clients is an important task to improve the local model performance first, then propagating these improvements into the global model after new training rounds in a privacy-preserving manner. We have engineered an intuitive process for each local client to contribute to the adding of new samples on top of their local data (Figure \ref{active_l}). The first step is to add a new data point that will remain on the local site, this allows the user to add their private data, questions, and correct answers. The incremental learning module will process and transform the private data locally and generate training points to be used in the local training. This process enables collaborative data generation between organizations in a private way in order to strengthen data protection and avoid unnecessary sharing within the alliance. In addition, a database layer is used to store user queries and global model predictions as feedback to enhance and improve the performance for further usage.

\IncMargin{1em}
\begin{algorithm}
\SetKwData{Left}{left}\SetKwData{This}{this}\SetKwData{Up}{up}
\SetKwFunction{Union}{Union}\SetKwFunction{FindCompress}{FindCompress}
\SetKwInOut{Input}{input}\SetKwInOut{Output}{output}
\Input{$D_i$}
\Output{$M(W_t)$}
\BlankLine
\emph{$W_0\leftarrow \emptyset$ }\;
\For{$t\leftarrow 1$ \KwTo $r$}{
\emph{$S_t\leftarrow \Gamma(k, B)$}\;
\For{client $k \in S_t$}{\label{forins}
\emph{In parallel}\;
$W_{t+1}^k \leftarrow ClientUpdate(k, W_t)$\;
$W_{t+1} \leftarrow \sum_{k=1}^{k} \frac{n_k}{n} W_{t+1}^k$\
}
$W_t \leftarrow (W_{t-1} + (W_{t}-W_{t-1})/t)$\
}
\Return $M(W_t)$
\caption{Incremental FedAvg algorithm.  $\Gamma$: Participants, k: number of clients, r: number of rounds, $D_i$: Datasets, $W_i$: model weights and $M$: Global models}
\label{fedavg}
\end{algorithm}\DecMargin{1em}


\begin{algorithm}[t!]
\DontPrintSemicolon
\SetAlgoLined
\KwOut{\textbf{$W_t$}}

\SetKwFunction{client}{ClientUpdate} // Run on client k\\
\SetKwProg{Fn}{Function}{:}{}
\Fn{\client{$k, W_t$}}{
$B \leftarrow (split~P_k~into~batches~of~size~B$)\\
\For{$local~epoch~e_i \in 1, \dots e$}{
\For{batch~b $\in$ B}{
$W_t \leftarrow W_t - \eta \nabla l(W_t,b)$\;
}
}
\KwRet{$W_t$}\;
}
\caption{Local client update, k: number of clients, B: sample size, e: number of local epochs, 
and $\eta$ is the learning rate }
\label{client-update}
\end{algorithm}

\section{Experiment and results}
\label{sec:experiments}
We conduct a validation experiment on the SQUAD 1.0 dataset, which contains 100.000+ question-answer pairs on 500+ articles. To ensure collaborative training, we randomly select and split data over 5 clients with 20\% for validation dataset for all clients. We used the BERT base as an encoder to build our model and the implementations are based on the public TensorFlow implementation from Keras. For the fine-tuning in our task, we set the initial learning rate to 5e-5. The batch size is set to 8. The maximum number of epochs is set to 1. Texts are tokenized using Wordpieces \citep{wu2016googles} with a maximum length of 384. More configuration is shown in Table \ref{training-config}.

\begin{table}
\centering
  \caption{Federated training configuration}
  \label{training-config}
  \begin{tabular}{lllll}
  \hline\noalign{\smallskip}
    Rounds & Nbr of clients & Update size &  Nbr of parameters\\
    \noalign{\smallskip}\hline\noalign{\smallskip}
	5 & 5 & 400 MB & 109.483.776\\ 
\noalign{\smallskip}\hline
\end{tabular}
\end{table}

\subsection{Evaluation}
For the evaluation, we used exact match (EM) and F1 score, the main metrics commonly used for question answering systems. These metrics are computed on individual (question, answer) pairs. In case of multiple correct answers for a given question, the maximum score over all possible correct answers is computed. In the EM metric, For each pair (question, answer), if the characters of the model's prediction exactly match the characters of (one of) the True Answer(s), EM = 1, otherwise EM = 0.

The Accuracy represents the percentage of the questions that an MRC system accurately answers. Each question corresponds to one correct answer. For the span prediction task, the accuracy is the same as Exact Match and can be computed by the formula \ref{EM} as follows:

\begin{equation}
Accuracy = EM = \frac{Number~of~correct~answers}{Number~of~questions}
\label{EM}
\end{equation}

The precision represents the percentage of token overlap between the tokens in the correct answer and the tokens in the predicted answer, while the recall is the percentage of tokens in a correct answer that have been correctly predicted in a question. The True Positive (TP) denotes the same tokens between the predicted answer and the correct answer, the False Positive (FP) denotes the tokens which are not in the correct answer but the predicted answer while the False Negative (FN) presents the tokens which are not in the predicted answer but the correct answer. Precision and Recall can be computed by the formulas \ref{precision} and \ref{recall} as follows:

\begin{equation}
Precision = \frac{N(TP)}{N(TP) + N(FP)}
\label{precision}
\end{equation}

\begin{equation}
Recall = \frac{N(TP)}{N(TP) + N(FN)}
\label{recall}
\end{equation}

The F1 score is a measure of a test's accuracy. It is the weighted average between precision and recall. The formula for this score is given in \ref{F1}. In our case, it's computed over the individual words in the prediction against those in the True Answer. The number of shared words between the prediction and the truth is the basis of the F1 score.

\begin{equation}
F1~score = 2\times \frac{(Precision\times Recall)}{(Precision + Recall)}
\label{F1}
\end{equation}

\begin{figure}[!h]
\centering
\includegraphics[width=8cm]{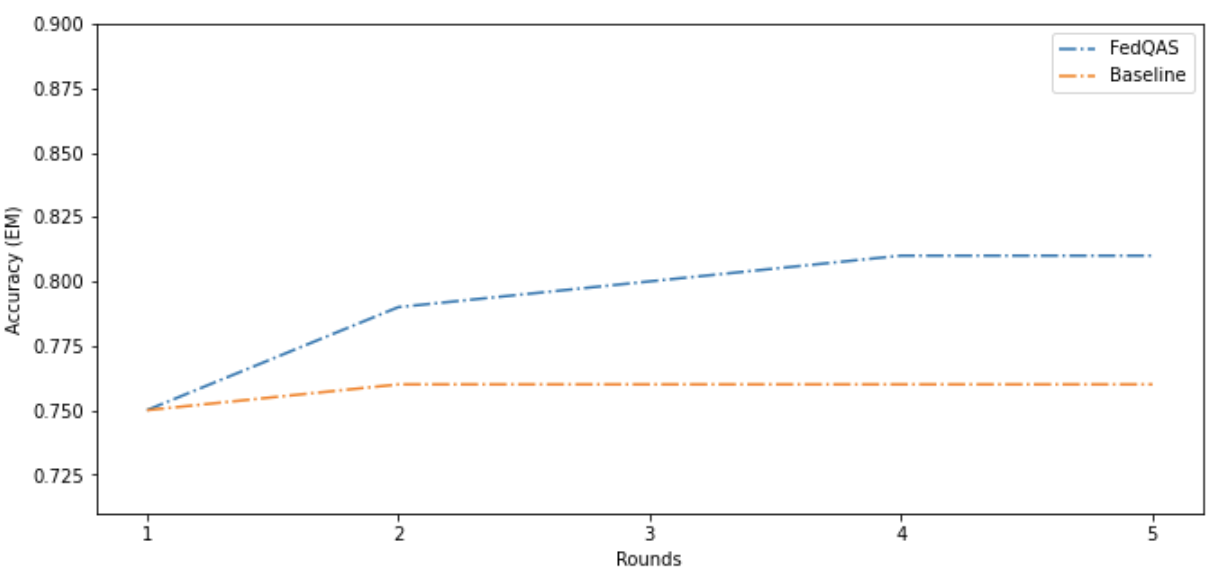}
\caption{Convergence of accuracy (Exact Match) on the SQUAD dataset with 1 combiner, 5 clients and 5 rounds}
\label{res-em}
\centering
\end{figure}

\begin{figure}[!h]
\centering
\includegraphics[width=8cm]{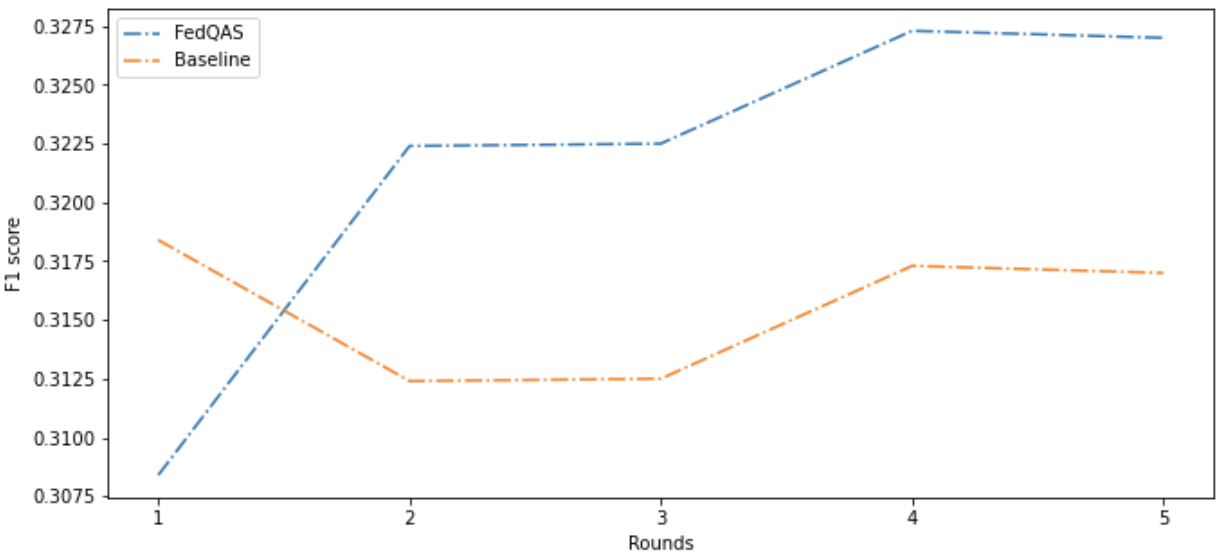}
\caption{Convergence of F1 score on SQUAD dataset with 1 combiner, 5 clients and 5 rounds}
\label{res-f1}
\centering
\end{figure}

To demonstrate the benefit of FedQAS, we partitioned the SQUAD dataset into 5 equal chunks, so that each client has "20\%" of the total dataset. We then compare the federated scenario to centralized model training. Table \ref{results} lists the available metrics for different training rounds of the global model. Our implemented model baselines show similar EM and F1 scores with the global model during the first rounds and slightly outperform the baseline with respect to data privacy and enabling knowledge sharing across participants. Overall, the result shows that question-answering in federated learning settings performs well compared to centralized settings, see Figure \ref{res-em} for the convergence of accuracy (EM) and Figure \ref{res-f1} for the convergence of F1.\\

\begin{table}[h!]
  \centering
  \caption{Comparisons with equivalent parameters on the validation set of SQuAD1.0}
  \label{results}
  \begin{tabular}{lll}
  \hline\noalign{\smallskip}
    Model  & F1 score & Accuracy (EM) \\
    \noalign{\smallskip}\hline\noalign{\smallskip}
	Baseline  	& 0.31 & 0.75\\ 
	\noalign{\smallskip}\hline\noalign{\smallskip}
	FedQAS  & 0.33 & 0.81 \\
\noalign{\smallskip}\hline
\end{tabular}
\end{table}

\begin{figure}
\centering
\includegraphics[width=8cm]{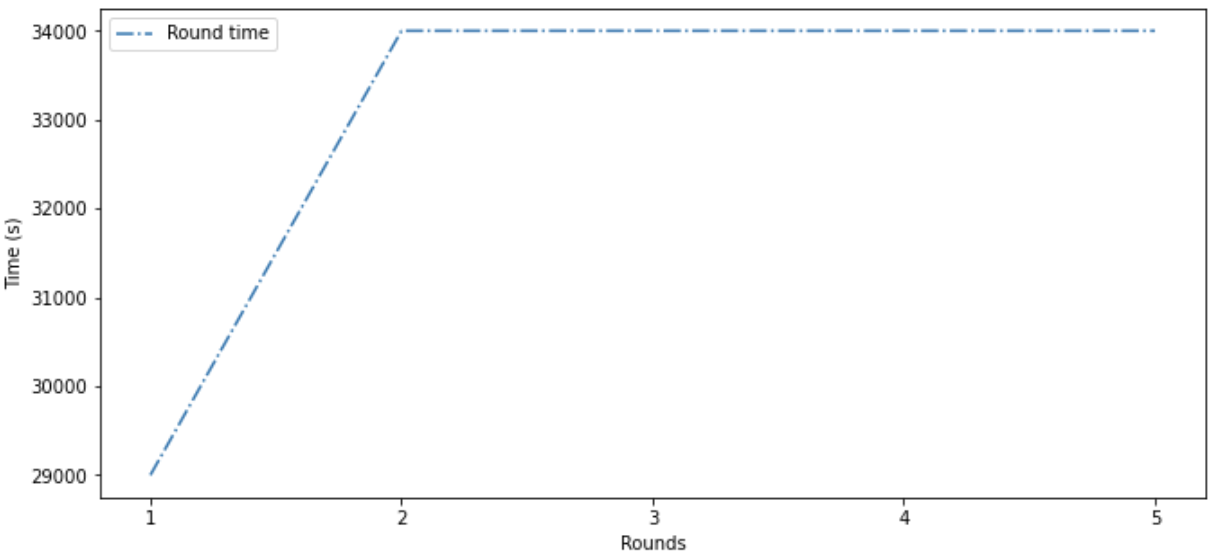}
\caption{Round times for global model training (FEDn)}
\label{fig:round-time}
\centering
\end{figure}

In terms of resources, the result proves the fact that model architecture affects client training time and combiner round time. Hence, training a large model (±400MB) in a centralized way requires more resources than the federated setting. For demonstration, we consider a FEDn network consisting of a single, high-powered combiner (8 VCPU, 32GB RAM) with connected clients (8 VCPU, 32GB RAM) instances in SSC (SNIC Science Cloud \citep{8109140}) and measure the average round time over 5 global rounds. Figure \ref{fig:round-time} shows round time for global model training.

To gain an intuitive observation of the predictions, we give a prediction example on SQuAD1.0 from both the baseline and federated model in Table \ref{prediction}, which shows that FedQAS works better at answering the question on a given passage. Hence, the proposed approach has contributed overall to a better understanding of QA, preserving data privacy, and contributed to low-cost training as well as a collaborative question answering system task using very large models.\\

\begin{table*}
  \caption{Comparison of answer prediction on test data}
  \label{prediction}
  \begin{tabular}{l}
  \hline
    \textbf{Title:} Project Apollo\\
    \noalign{\smallskip}\hline\noalign{\smallskip}
    \textbf{Passage:} The Apollo program, also known as Project Apollo, was the third United States human spaceflight program \\ carried out by the National Aeronautics and Space Administration(NASA), which accomplished landing the\\ first humans on the Moon from 1969 to 1972. First conceived during Dwight D. Eisenhower's administration \\ as a three-man spacecraft to follow the one-man Project Mercury which put the first Americans in space,\\ Apollo was later dedicated to President John F. Kennedy's national goal of landing a man on the Moon and\\ returning him safely to the Earth by the end of the 1960s, which he proposed in a May 25, 1961,\\ address to Congress. Project Mercury was followed by the two-man Project Gemini. The first manned flight \\ of Apollo was in 1968. Apollo ran from 1961 to 1972, and was supported by the two man Gemini program\\ which ran concurrently with it from 1962 to 1966...\\
    \noalign{\smallskip}\hline\noalign{\smallskip}
	\textbf{Question 1:} How long did Project Apollo run? \\ 
	\noalign{\smallskip}\hline\noalign{\smallskip}
	\textbf{Gold answer (human):} 1961 to 1972 \\
	\textbf{Google search engine answer:} see Figure \ref{fig1:mrc}\\
	\textbf{Baseline model answer:} 1961 to 1972 \\ 
	\textbf{FedQAS answer:} 1961 to 1972\\
	
	\noalign{\smallskip}\hline\noalign{\smallskip}
	\textbf{Question 2:} What program was created to carry out these projects and missions? \\ 
	\noalign{\smallskip}\hline\noalign{\smallskip}
	\textbf{Gold answer (human):} Apollo program \\
	\textbf{Baseline model answer:} National Aeronautics and Space Administration \\ 
	\textbf{FedQAS answer:} Apollo program\\
	
	\noalign{\smallskip}\hline\noalign{\smallskip}
	\textbf{Question 3:} What year did the first manned Apollo flight occur? \\ 
	\noalign{\smallskip}\hline\noalign{\smallskip}
	\textbf{Gold answer (human):} 1968 \\
	\textbf{Baseline model answer:} 1968 \\ 
	\textbf{FedQAS answer:} 1968\\
	
	\noalign{\smallskip}\hline\noalign{\smallskip}
	\textbf{Question 4:} What President is credited with the original notion of putting Americans in space? \\ 
	\noalign{\smallskip}\hline\noalign{\smallskip}
	\textbf{Gold answer (human):} John F. Kennedy \\
	\textbf{Baseline model answer:} John F. Kennedy \\ 
	\textbf{FedQAS answer:} John F. Kennedy\\
	
	\noalign{\smallskip}\hline\noalign{\smallskip}
	\textbf{Question 5:} Who did the U.S. collaborate with on an Earth orbit mission in 1975? \\ 
	\noalign{\smallskip}\hline\noalign{\smallskip}
	\textbf{Gold answer (human):} Soviet Union \\
	\textbf{Baseline model answer:} Soviet Union \\ 
	\textbf{FedQAS answer:} Soviet Union\\
	
	\noalign{\smallskip}\hline\noalign{\smallskip}
	\textbf{Question 6:} How long did Project Apollo run? \\ 
	\noalign{\smallskip}\hline\noalign{\smallskip}
	\textbf{Gold answer (human):} 1962 to 1966 \\
	\textbf{Baseline model answer:} 1961 to 1972, and was supported by the two man Gemini program which ran 1966 \\ 
	\textbf{FedQAS answer:} 1962 to 1966\\
	
	\noalign{\smallskip}\hline\noalign{\smallskip}
	\textbf{Question 7:} What program helped develop space travel techniques that Project Apollo used? \\ 
	\noalign{\smallskip}\hline\noalign{\smallskip}
	\textbf{Gold answer (human):} Gemini \\
	\textbf{Baseline model answer:} Gemini \\ 
	\textbf{FedQAS answer:} Gemini \\

	\noalign{\smallskip}\hline\noalign{\smallskip}
	\textbf{Question 8:} What space station supported three manned missions in 1973-1974? \\ 
	\noalign{\smallskip}\hline\noalign{\smallskip}
	\textbf{Gold answer (human):} Skylab \\
	\textbf{Baseline model answer:} Skylab \\ 
	\textbf{FedQAS answer:} Skylab \\
	
\hline
\end{tabular}
\end{table*}

\subsection{Implementation and demo environment}
Designing and developing Question Answering in a privacy-aware manner is not a trivial task. It requires design strategies to comply with data governance and privacy regulations. Several third-party frameworks have been proposed for federated learning; providing open source building blocks that help to collaborate in training machine learning models. In the present QAS application such a framework need to provide scalability, large models training, and production-grade features such as robustness to failure. Based on these requirements, we chose to design and develop our proposed FedQAS system on top of private data using the FEDn framework \citep{ekmefjord2021scalable}. FEDn is an open-source, modular, and model agnostic framework for federated machine learning. We developed interactive and user-friendly interfaces using Flask framework \footnote{https://flask.palletsprojects.com} which make it easy for a third party to contribute to data annotation and then participate in training global models directly from their location site. The proposed FedQAS has the following features:

\begin{itemize}
    \item Privacy-preserving: sharing only model parameters with a central server (cloud) and keeping data private on client side,
    \item Incremental learning: improving the global models by attaching more clients and adding new data points,
    \item Robust: robust enough to deal with natural language tasks (e.g., question answering, chatbot, etc.) and large models in a geographically distributed manner,
    \item Multilingual: language agnostic, can be trained on any language,
    \item Standalone: multiple platforms (i.e., guarantee for low disk and memory footprint). It can be run production-grade on a standard laptop having two cores and 2GB of RAM,
    \item Accuracy and F1 score: achieve competitive performance compared with centralized training and the used baseline model (see experiment and evaluation section).
\end{itemize}

\vspace{0.2cm}
The proposed FedQAS is composed of three main components: FEDn for collaborative training, MongoDB \citep{mongodb} as a NoSQL \citep{10.1145/2095536.2095583} database and question answering UI for prediction and local incremental learning. The system is interactive, scalable, suitable for secure collaborative training and data privacy-preserving, and can be used both in the cloud, on edge nodes and in a standalone mode. It is accessible from different platforms to engage a wide range of users, and it is also optimized for both desktop and mobile. The source code is publicly available on Github via this link \url{https://github.com/aitmlouk/FEDn-client-FedQAS-tf.git}.\\

To summarise, FedQAS is, to the best of our knowledge, the only approach for question answering that supports data privacy and knowledge sharing. Its main value is to provide an environment to quickly ensure data privacy and low-cost training by collaborative training. Nearly every deep learning application can benefit from data privacy and knowledge sharing across a different client in a federated learning setting. The transformer model used in FedQAS can be improved by tuning parameters and using transfer learning for new pre-trained models (GPT-2 \citep{radford2019language}, GPT-3 \citep{brown2020language}, etc.)

\section{Conclusion}
\label{sec:conclusion}
In this paper, we have proposed FedQAS, a high-quality question answering approach, to address the data-sharing issue in machine learning. Validation experiments for FedQAS was implemented based on 5 rounds of training with a transformer neural network. The system consists of several components including the private data pipeline, collaborative training and private incremental learning. Experiments on the SQUAD dataset using the transformer architecture demonstrate that our FedQAS significantly outperforms the baseline model performances, protecting data privacy and sharing knowledge within an alliance. The proposed FedQAS allows collaborators (collaborative training participants) to have overall control of their sensitive and private data while collaboratively training question answering models. The integration of federated learning within machine reading comprehension provides a sustainable solution by preserving data privacy and ensuring low-cost training. We conclude that the application of FL to NLP tasks such as question answering can contribute to solving the problem that arises when using machine learning in the context of data protection and privacy. The system actively supports end-users in joining training and improving the performance through incremental learning on a various range of local clients. In future work, we aim to extend the FedQAS to cover more question answering datasets and allow federated learning for large private documents.

\section*{Availability}
FedQAS is publicly available under the Apache2 license at \url{https://github.com/aitmlouk/FEDn-client-FedQAS-tf}.

\section*{Acknowledgement}
This research is funded by the Swedish eSSENCE strategic collaboraion on eScience. Authors also would like to acknowledge SNIC for cloud resources.

\bibliographystyle{unsrtnat}
\bibliography{fedqas}

\end{document}